\documentclass[10pt,twocolumn,letterpaper]{article}

\usepackage{iccv}
\usepackage{times}
\usepackage{epsfig}
\usepackage{graphicx}   
\usepackage{booktabs}   
\usepackage{amsmath}
\usepackage{amssymb}
\usepackage{cite}
\usepackage[dvipsnames]{xcolor}
\usepackage{bm}
\usepackage{graphicx}
\usepackage{array}
\usepackage{multirow}
\usepackage{booktabs}
\usepackage{tikz}
\usepackage{float} 
\usepackage{caption} 
\usepackage{textcomp}
\usepackage{arydshln}
\usepackage{subcaption}
\usepackage{etoolbox}

\definecolor{citecolor}{HTML}{0071bc}
\usepackage[pagebackref=false,breaklinks=true,letterpaper=true,colorlinks,citecolor=citecolor,bookmarks=false]{hyperref}

\usepackage[titletoc,title]{appendix}

\usepackage{algorithm}
\usepackage{algorithmic}

\usepackage{graphicx}
\usepackage{caption}

\iccvfinalcopy 

\ificcvfinal\fi

\begin{document}
\makeatletter
\let\@oldmaketitle\@maketitle
\renewcommand{\@maketitle}{\@oldmaketitle
\begin{minipage}{\textwidth}
\vspace{-0.8cm}
\centering
\includegraphics[width=1\linewidth]{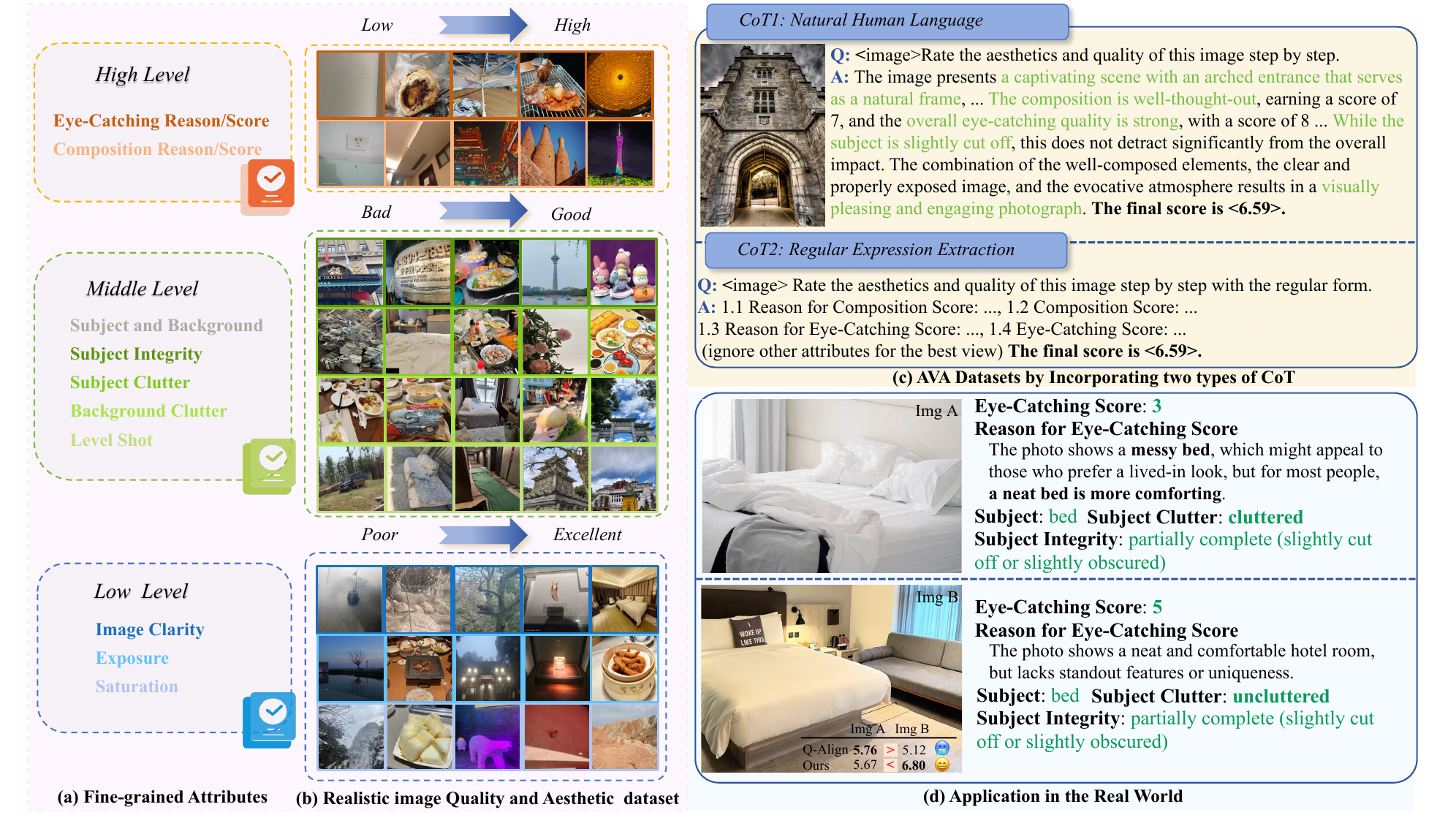}
    \vspace{-6.5mm}
\captionof{figure}{\textbf{Real}istic image \textbf{Q}uality and \textbf{A}esthetic (RealQA) dataset (b), including 10 rich fine-grained attributes (a). Based on these attributes, we can (1) reconstruct public datasets in a CoT manner and (2) directly apply the fine-grained attributes and composite scores for real-world applications. We show two kinds of CoT forms on the AVA dataset in (c). And we show the real-world applications in (d), where Q-Align \cite{qalign} trained on the AVA dataset \cite{ava} assigns an inappropriate rank (scores scale to 1–10 for fairness), while the model trained on the RealQA dataset gives a correct rank with the rich fine-grained attributes. Here, we show only part of the predicted fine-grained attributes for clearer viewing.}
    \label{fig:dataset}
    \vspace{+0.3cm}
\end{minipage}}

\makeatother
\vspace{1cm} 

\title{Next Token Is Enough: Realistic Image Quality and Aesthetic Scoring with Multimodal Large Language Model}
\author{Mingxing Li$^{1*}$, Rui Wang$^{2*}$, Lei Sun$^1$, Yancheng Bai$^1$, Xiangxiang Chu$^1$ \\ $^{1}$ Amap, Alibaba Group \ \  $^{2}$ BUPT 
\\ {\tt\small \{limingxing.lmx, ally.sl\}@alibaba-inc.com}
}

\maketitle
\ificcvfinal\thispagestyle{empty}\fi

\begin{abstract}
\vspace{-0.3cm}
The rapid expansion of mobile internet has resulted in a substantial increase in user-generated content (UGC) images, thereby making the thorough assessment of UGC images both urgent and essential. Recently, multimodal large language models (MLLMs) have shown great potential in image quality assessment (IQA) and image aesthetic assessment (IAA). Despite this progress, effectively scoring the quality and aesthetics of UGC images still faces two main challenges: 1) A single score is inadequate to capture the hierarchical human perception.  2) How to use MLLMs to output numerical scores, such as mean opinion scores (MOS), remains an open question. To address these challenges, we introduce a novel dataset, named \textbf{Real}istic image \textbf{Q}uality and \textbf{A}esthetic (RealQA), including \textbf{14,715} UGC images, each of which is annotated with \textbf{10} fine-grained attributes. These attributes span three levels: low level (e.g., image clarity), middle level (e.g., subject integrity) and high level (e.g., composition). Besides, we conduct a series of in-depth and comprehensive investigations into how to effectively predict numerical scores using MLLMs. Surprisingly, by predicting just two extra significant digits, the next token paradigm can achieve SOTA performance. Furthermore, with the help of chain of thought (CoT) \cite{cot} combined with the learnt fine-grained attributes, the proposed method can outperform SOTA methods on \textbf{five} public datasets for IQA and IAA with superior interpretability and show strong zero-shot generalization for video quality assessment (VQA). The code and dataset will be released.\end{abstract}

\section{Introduction}
The widespread adoption of mobile internet enables users to effortlessly upload images, leading to the large-scale generation of UGC images. Scoring UGC images in a way which closely aligns with human perception becomes increasingly important in real-world applications.

Recently, numerous advancements \cite{Q-bench, qalign, wu2025comprehensive, zhu2024adaptive, zhu20242afc, you2024descriptive} have been made in leveraging MLLMs for IQA and IAA, due to their exceptional capabilities in visual and linguistic understanding. Despite this progress, effectively scoring the quality and aesthetics of the UGC images still faces two main challenges: \textbf{1) A single score (e.g., MOS) is inadequate to capture the hierarchical human perception.} Simply learning one score restricts the capability of MLLMs to capture the underlying rationale behind judgments, thereby limiting the effective alignment with human perception \cite{vila, vera2022understanding, zhou2024uniaa}. We believe that incorporating fine-grained attribute-level content can promote more robust consistency with human perception and better interpretability.
\textbf{2) How to utilize MLLMs to predict the final score (e.g., MOS) is still an open question.} Current approaches typically employ either the pre-defined textual labels (e.g., ``poor", ``bad") \cite{Q-bench, qalign} to associate with the limited discrete scores or direct regression to a numerical score from hidden layers \cite{he2024videoscore}. Nevertheless, the pre-defined textual labels pose challenges for further refinements. For example, if we define a word between ``poor” and ``bad”, it is hard to find it accurately in the predefined vocabulary. Meanwhile, scores predicted by the regression method cannot be seamlessly integrated with the next-token prediction paradigm. This motivates us to ask: \textit{Is it possible for MLLMs to directly predict the final score numerically?} 

To address these challenges, we first propose a novel dataset, named \textbf{Real}istic image \textbf{Q}uality and \textbf{A}esthetic (RealQA) dataset. To align with the human perception, as shown in Fig.\ref{fig:dataset} (a), we decompose human perception into three perspectives: low-level attributes, middle-level attributes and high-level attributes, which includes \textbf{10 fine-grained human perceptual attributes} in total. Specifically, the high-level attributes offer a more comprehensive assessment of an image, such as composition and the degree of the eye-catching. The middle-level attributes differentiate the layering between foreground and background and emphasize the expression of neatness and integrity. The low-level attributes pertain to the fundamental quality of images, including clarity, exposure, and saturation. The RealQA dataset collects \textbf{14,715} images from AutoNavi, which are taken in various industries, including tourist attractions, restaurants, hotels, leisure and entertainment venues, and other user-active areas. To ensure the application for the real-world UGC image scenarios, we collect the feedback in the real application to determine the weights of various attributes by partial least squares. These weights are aggregated into a composite score, facilitating the comprehensive evaluation of the fine-grained attributes. Although the composite score does not follow the MOS collection, it relies on the online results to better fit the real applications.

\begin{figure}[!t]
\centering
\includegraphics[width=.95\linewidth]{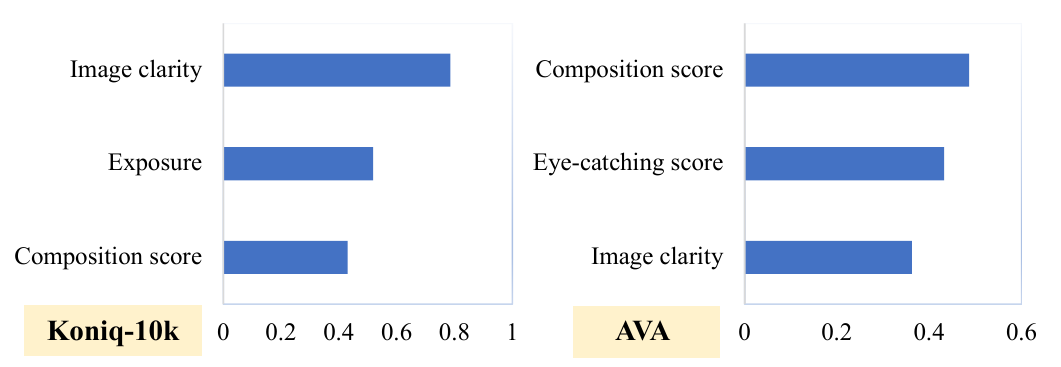}
\caption{Top 3 Pearson correlation coefficients between \textbf{predicted attributes} and \textbf{MOS} on the IQA (Koniq-10k \cite{hosu2020koniq}) and IAA (AVA \cite{ava}) dataset separately. Some attributes already have a high correlation with the MOS for IQA and IAA, which can be utilized to improve the ability of MLLMs to score the image quality and aesthetics.
}
\vspace{-0.2cm}
\label{fig:fig1}
\end{figure}

Second, since a numerical score is generally composed of multiple tokens in MLLMs (e.g., for 3.99, the token of Qwen2-VL \cite{wang2024qwen2} is ``3", ``.", ``9" and ``9" ), it is not a simple one-time token classification problem. Based on the above observation, we conduct a series of in-depth and comprehensive research on how to use MLLMs to directly predict scores numerically. Interestingly, we find several conclusions:
a) For simple numerical sorting, even MLLMs of similar sizes vary greatly in their performance.
b) Training with the next token paradigm, predicting two extra significant digits can greatly improve model performance compared to directly predicting the integer.
c) We propose the Numerical Continuity Metric (NCM) and its variant NCM$^{*}$, which verify that well-trained MLLMs understand numerical scores as a whole rather than memorizing tokens in different positions.
Although previous methods \cite{he2024videoscore, qalign} show inferior performance when MLLMs directly predict the final numerical score. In this paper, we argue that it is entirely feasible to employ MLLMs for accurate numerical score prediction within the next token prediction paradigm.

Furthermore, relying on the fine-grained attributes labeled on the RealQA dataset, we can utilize a cold-start strategy to empower MLLMs with the capability to extract these attributes. Then, naturally, we can self-label the attributes of images in the public datasets (e.g. Koniq-10k and AVA). As shown in Fig.\ref{fig:fig1}, some of the attributes already have a high correlation with the MOS on the IQA and IAA datasets. Thus, we organize the attributes and the final scores using CoT for the public datasets. We retrain the final model to simultaneously predict the attributes and the final score. With the help of CoT, the proposed method outperforms the SOTA method on \textbf{five} public IQA and IAA datasets. For example, the proposed method surpasses Q-Align by achieving a higher PLCC by $\textbf{+1.8\%}$ on the Koniq-10k dataset and $\textbf{+4.5\%}$ on the cross-domain LIVE Challenge \cite{livec} dataset.  Furthermore, we demonstrate strong generalization on the video quality assessment (VQA) dataset KoNViD \cite{konvid}. The model trained only with image data has a $\textbf{+36.4\%}$ improvement in SRCC.

Our core contributions can be summarized as three-fold:
\vspace{-0.5cm}
\begin{itemize}
\item We introduce a novel UGC assessment dataset, named RealQA, a collection of \textbf{14,715} UGC images. Each image is annotated with \textbf{10} fine-grained human-perceptual attributes, reflecting the common perception about image quality and aesthetics.
\vspace{-0.1cm}
\item We conduct a series of in-depth research of the next token paradigm on how to use MLLMs to directly predict numerical scores. We find that based on the model training extensively on large-scale data, directly predicting the numerical scores with two extra significant digits can achieve superior performance.
\vspace{-0.1cm}
\item By leveraging CoT to integrate fine-grained attributes, the proposed method surpasses SOTA performance on \textbf{five} IQA and IAA benchmarks and demonstrates strong zero-shot generalization for the VQA task.
\end{itemize}
\vspace{-0.1cm}

\section{Related Work}
\subsection{Connection between IQA and IAA}

In recent years, IQA has gradually shifted towards deep learning-based full-reference or no-reference methods \cite{cheon2021perceptual, GRIDS, TRIQ, qalign, clip_iqaplus, AVMHAR,chen2024promptiqa, chen2024seagull}. By integrating multi-scale feature fusion \cite{cvpr_liqe}, these studies have improved performance by decomposing quality attributes into quantifiable dimensions such as synthetic degradations (e.g., Gaussian noise) \cite{live, lin2019kadid} and real-world degradations (e.g., sensor noise) \cite{hosu2020koniq, livec, spaq}. IAA has experienced an evolution from artificial features (e.g., color histogram \cite{datta2006studying}) to data-driven models \cite{PARA, AesCLIP}. 
Although both IQA and IAA aim to simulate human subjective perception, most existing methods treat them independently, overlooking their complementary nature at the visual perception level. For example, aesthetic features like color balance may significantly affect quality ratings, while quality defects like noise may reduce aesthetic appeal \cite{zhou2024uniqa, zhu2024attribute}. Recent studies on AI-Generated Content (AIGC) images indicate that integrating IQA and IAA by jointly learning shared representations can improve model ability to comprehend image perception \cite{MPS, yuan2024pkuaigiqa4k}. AGIN \cite{AGIN} enhances naturalness attributes (e.g., brightness) and evaluates images based on presence, color, and layout, addressing the limitations of sparse single-task data and noisy annotations. The proposed RealQA dataset integrates IQA and IAA via the fine-grained attributes, creating a more effective and user-centric assessment system.

\subsection{Quality and Aesthetic Scoring with MLLMs}
Recently, MLLMs \cite{gpt4o, guo2024llava, chen2024internvl, wang2024qwen2, chen2024sharegpt4v} seamlessly integrate visual and linguistic information, exhibiting significant potential. Recently, there has been significant progress in research on scoring for IQA and IAA utilizing MLLMs \cite{huang2024visualcritic, huang2024aesexpert, mi2024clif, CD-Reasoning}.
DepictQA \cite{you2312depicting} uses MLLMs to generate reasoning description language for IQA to explain the image quality, but does not output the final score. When utilizing MLLMs, how to model numerical scores is an open problem. Q-Bench \cite{Q-bench} and Q-Align \cite{qalign} suggest using the discrete text-defined levels. Among them, the most representative method is Q-Align \cite{qalign}.  Q-Align suggests predicting the discrete text-defined levels (``excellent", ``good", ``fair", ``poor", and ``bad"), which are more suitable for MLLMs to predict. In implementation, these discrete text-defined levels represent different intervals of MOS.  However, although the text-defined levels simulate human language habits, further refinement remains challenging. For example, if we define a word between ``poor” and ``bad”, it is difficult to find it accurately in the predefined vocabulary. Additionally, Videoscore \cite{he2024videoscore} predicts the numerical scores by the regression from the linear layer. However, the regression result cannot be naturally output together with the next token prediction paradigm and lacks interpretability. Thus, we conduct a series of in-depth investigations into how to effectively predict numerical scores using MLLMs.

\section{Method}
In the following sections, we thoroughly describe the detailed procedures involved in creating the RealQA dataset in Sec.\ref{sec:realqa}. Then, Sec.\ref{sec:stage2} describes numerical ambiguity, the proposed NCM and its variant NCM$^{*}$. Next, Sec.\ref{sec:training_pipeline} outlines the comprehensive training recipe, where we present the fine-grained attributes training and CoT training with multiple forms. Finally, we show the model architecture.
\begin{figure*}[!t]
\centering
\includegraphics[width=0.95\textwidth]{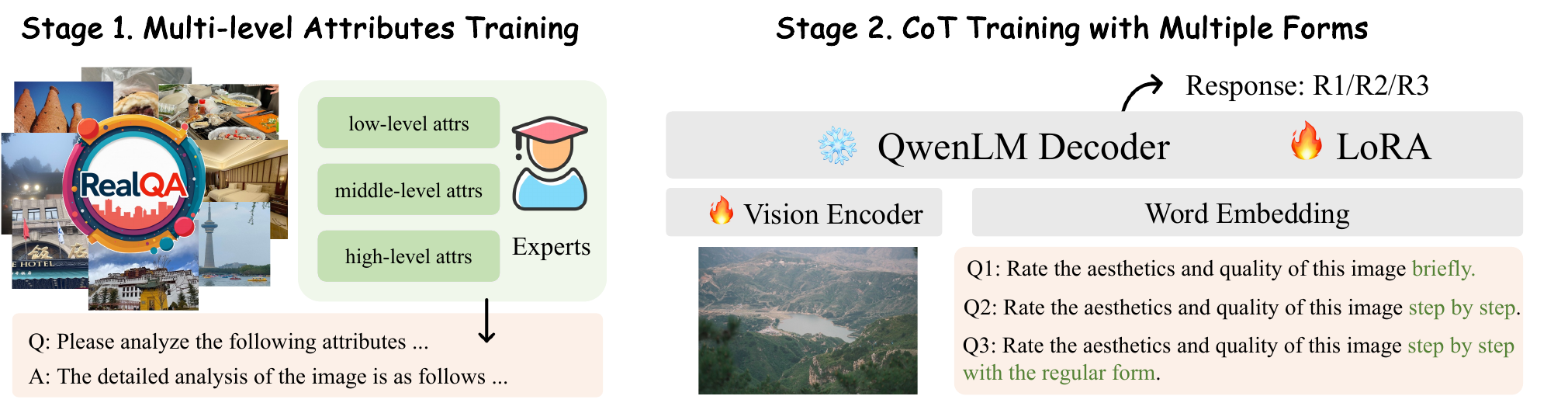}
\caption{Training pipeline. In stage 1, we utilize a cold-start strategy to empower MLLMs with the capability to extract the fine-grained attributes. Then, we self-label the fine-grained attributes of images in public datasets. In stage 2, we retrain the model by combining the fine-grained attributes and numerical scores for public datasets in the CoT manner. }
\label{fig:pipeline}
\vspace{-0.2cm}
\end{figure*}

\subsection{Realistic Image Quality and Aesthetic Dataset}\label{sec:realqa}
\paragraph{\noindent\textbf{Data Curation.}}
Adequate image quality and aesthetic scoring generally depend on a wide range of image sources. To reflect this variability, we collect 14,715 UGC images from AutoNavi, including tourist attractions, restaurants, hotels and other user-active areas. These images undergo processing steps, including subtitle and watermark filtering and manual image stitching filtering. After the filtering processes, we divide the dataset into a training set containing 13,712 images and a test set containing 1,003 images.

\paragraph{\noindent\textbf{Image Attributes Selection.}}
Human perception of image quality and aesthetics relies on multiple attributes. As shown in Fig.\ref{fig:dataset} (a), we meticulously categorize image quality and aesthetics into three perspectives: high-level attributes, middle-level attributes, and low-level attributes. 
\textbf{From the high-level perspective}, the eye-catching score of the image describes how attractive the content of the image is. The composition score represents how well the arrangement and organization of elements within a work create a harmonious and aesthetically pleasing whole. \textbf{From the middle-level perspective}, we first identify the subject and the background. Based on the subject and the background, we choose the subject integrity, subject clutter and background clutter. The subject integrity describes how complete the subject is. The subject clutter describes the degree of clutter of the subject, which is usually useful in an image related to food, hotels, a group of subjects and so on. Similarly, the background clutter examines the visual noise in the background that might distract from or compete with the subject, influencing the overall coherence and balance of the image. Besides, the level shot determines whether the image is taken straight or non-horizontally. \textbf{From the low-level perspective}, we select image clarity, exposure and saturation, which pertain to the fundamental quality of images.
\vspace{-0.3cm} 
\paragraph{\noindent\textbf{Annotation.}}
For high-level attributes, we assess a score ranging from 1 to 10 and provide appropriate reasons to support the evaluations. For attributes at other levels, we employ multi-tiered classifications. For example, subject and background clutter are categorized as cluttered, moderately cluttered and uncluttered. See appendix B.2 for all attribute details). However, annotating fine-grained attributes is challenging. Initially, we employ professionally trained annotators to annotate the fine-grained attributes. While there is significant variability in the annotation of fine-grained attributes. To address this issue, we assign each tag to some specific annotators. Nevertheless, as individual annotators handle more items, their annotations tend to become more neutral and conservative. Ultimately, using prompt engineering, we extensively adopt closed-source MLLMs (e.g., GPT-4o\cite{gpt4o}) to label the images. Subsequently, we instruct the annotators to refine the outputs generated by the closed-source MLLMs and correct the obvious errors when the prediction of MLLMs is wrong.

\subsection{Next Token Prediction Paradigm}\label{sec:stage2}
\begin{figure}[!t]
\centering
\includegraphics[width=.95\linewidth]{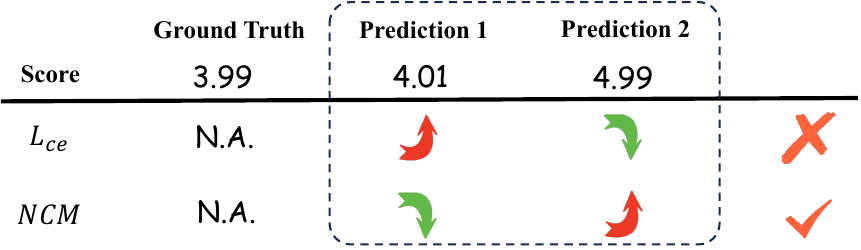}
\caption{The numerical ambiguity of the standard cross-entropy loss $\mathcal{L}_{ce}$ for predicting numerical scores. ``4.01" is a \textbf{more accurate} prediction than ``4.99", but the $\mathcal{L}_{ce}$ is \textbf{larger}. The proposed NCM and its variant NCM$^{*}$ transform the \textit{discrete tokens} into \textit{continuous expectations} to monitor the influence of the numerical ambiguity.
}
\label{fig:ncm_motivation}
\end{figure}

\paragraph{\noindent\textbf{Numerical Ambiguity.}}
When using MLLMs to predict numerical scores in the next token prediction paradigm, numerical scores consist of multiple tokens. For Qwen2-VL, the score 3.99 is typically represented by the tokens ``3", ``.", ``9", and ``9". During training with teacher-forcing \cite{lamb2016professor}, the cross-entropy loss $\mathcal{L}_{ce}$ can be formulated as follows:
\vspace{-0.1cm}
\begin{equation}
\mathcal{L}_{ce} = - \sum_{i=1}^{T} \log P(t_i | t_1, t_2, \ldots , t_{i-1}),
\label{eq:ce}
\end{equation}
where $t_i$ denotes the $i$-th token and $T$ denotes the sequence length. The cross-entropy loss only maximizes the probability of logits for the corresponding GT tokens, leaving other negative tokens unsupervised. For the numerical scores composed of multiple tokens, a natural question is \textbf{whether MLLMs memorizes the tokens at different positions or understands them as a whole.} For example, during inference, if the first digit is predicted incorrectly, it may cause numerical ambiguity, as shown in Fig.\ref{fig:ncm_motivation}. When the difference between the predicted number and GT number is larger, the cross-entropy loss is smaller.

\paragraph{\noindent\textbf{Numerical Continuity Metric.}}
To answer the question, we propose NCM and its variant NCM$^{*}$ to monitor the training stage with the supervision of the cross-entropy. However, the output tokens of direct autoregression are discrete. Therefore, we transform these discrete tokens into continuous mathematical expectations. Specifically, the GT numerical score $S_{GT} \in [0,10)$  can be formulated as:
\begin{equation}
\footnotesize
S_{GT} \in \left\{ \text{round}\left(\frac{k}{10^{M-1}}, M-1\right) \right.
\bigg| \left. k \in \mathbb{Z},\ 0 \leq k < 10^{M} \right\},
\end{equation}
where $M$ denotes the number of digits. We normalize scores from different datasets to this range (e.g., for score 3.9845, $S_{GT}$=4.0 when $M$=2 and $S_{GT}$=3.98 when $M$=3). The function $\text{round}(\frac{k}{10^{M-1}}, M-1)$ denotes rounds the number $\frac{k}{10^{M-1}}$ to $M-1$ decimal places. 
Let $\bm{z}^n \in \mathbb{R}^{M \times d}$ denotes the predicted logits corresponding to the digits, where $d$ is 10, due to only the digits 0 through 9 are considered \footnote{In Qwen2-VL, `0-9' and `.' are each a single token.}. We can calculate the mathematical expectation $\mathbb{E}[\bm{z}_i^n]$ corresponding to the $i$-th digit of the final numerical score $\bm{z}_i^n \in \mathbb{R}^d$, which is defined as follows:
\begin{equation}
    \mathbb{E}[\bm{z}_i^n] = \mathbf{v} \cdot \text{Softmax}(\bm{z}_i^n)  = \frac{\mathbf{v} \cdot e^{\bm{z}_i^n}}{\sum_{j=1}^de^{\bm{z}_j^n}}
\end{equation}
where $\mathbf{v} =  (0, 1, 2, \dots, 9)$ denotes the digits vector.
Furthermore, the expectation $\mathbb{E}[S_{pred}]$ of the predicted numerical score $S_{pred}$ can be formulated as follows: 
\begin{equation}
\begin{aligned}
    \mathbb{E}[S_{pred}] &= w_1 \mathbb{E}[\bm{z}_1^n] + w_2 \mathbb{E}[\bm{z}_2^n] + \cdots + w_M \mathbb{E}[{\bm{z}_M^n}] \\
    &= \bm{w} \cdot \frac{\mathbf{v} \cdot e^{\bm{z}^n}}{\sum_{j=1}^d e^{\bm{z}_j^n}}.
    \label{eq:s_pred}
\end{aligned}
\end{equation}
The numerical weight can be formulated as $\bm{w} = (1, 0.1, \cdots, 10^{1-M})$. Suppose the ground-truth score is $S_{GT} \in [0,10)$, the final NCM can be formulated as follows:
\begin{equation}
    \text{NCM} = \text{MSE}(\mathbb{E}[S_{pred}], S_{GT}).
\end{equation}

To further investigate the generalization of numerical scores by MLLMs, we propose a variant metric, $\text{NCM}{^{*}}$, which calculates the expectation \textit{when the GT digits are excluded}. The insight is that a well-trained MLLMs predict digits near the GT digits even when the first-ranked tokens (i.e., GT tokens) are excluded, due to adjacent tokens having high probabilities (e.g., for GT $t_i$=3, the adjacent tokens are 2 and 4). In contrast, poorly converged MLLMs tend to predict randomly distributed expectations.

The new expectation $\mathbb{E}[S_{pred}]^{*}$ of the predicted numerical score $S_{pred}$ can be formulated as follows:
\begin{equation}
    \mathbb{E}[S_{pred}]^{*} = \bm{w} \cdot \frac{ \mathbf{v} \cdot e^{\bm{m} \odot \bm{z}^n}}{\sum_{j=1}^d e^{\bm{z}_j^n}},
\end{equation}
where $\bm{m} \in \mathbb{R}^{d}$ denotes the binary mask. The position corresponding the GT digit is $-\infty$, and other positions are 1.
The final $\text{NCM}{^{*}}$ which ignores GT digits can be formulated as follows:
\begin{equation}
    \text{NCM}^{*} = \text{MSE}(\mathbb{E}[S_{pred}]^{*}, S_{GT}).
\end{equation}

\subsection{Training Recipe} \label{sec:training_pipeline}
\paragraph{\noindent\textbf{Training Pipeline.}}
As shown in Fig.\ref{fig:pipeline}, the training processing can be divided into two stages: fine-grained attributes training and CoT training with multiple forms. 
The first stage aims to utilize a cold-start strategy to empower MLLMs with the capability to extract the fine-grained attributes. The training conversation can refer to section \ref{sec:stage1_conversation}. The fine-grained attributes align with human perception and increase interpretability, which can further improve the understanding of MLLMs.
Then, we self-label the fine-grained attributes of images in public datasets. During the second stage, we retrain the model by combining the fine-grained attributes and the numerical scores for public datasets (e.g., AVA and KonIQ-10k) in the CoT manner. 
\paragraph{\noindent\textbf{Conversation Formats.}}\label{sec:stage1_conversation}
During the first stage, to accommodate diverse requirements and improve the capability of extracting the fine-grained attributes, we provide three specific conversation templates: 1) Q\&A for individual items (``Attributes"), 2) Q\&A for a single level (``Levels"), and 3) Q\&A for all items (``Mix"). ``Attributes” denotes that we predict an attribute in one conversation. ``Levels” denotes that we predict certain level attributes in one conversation. ``Mix” denotes that we predict all attributes in one conversation. See Appendix B.3 for specific templates. During the second stage, according to different purposes, we organize several conversation formats: 1) the direct answer to numerical scores (Q1-R1), 2) the CoT with a natural human language format (Q2-R2) and 3) the CoT with a format conducive to regular expression extraction (Q3-R3). We can get direct or CoT responses based on different questions. As shown in Fig.\ref{fig:dataset}, we show examples of the two CoT formats. Besides, the Q1-R1 format is as follows:

{\small
\noindent
\textit{\# User: Rate the aesthetics and quality of this image briefly.}

\noindent
\textit{\# Assistant: The score of this image is \textless$t_i$, $t_{i+1}$, ... , $t_{T'}$\textgreater.}
}

\begin{table*}[!t]
\centering
\resizebox{0.95\textwidth}{!}{
\begin{tabular}{c |c c|c c c c|c c c| c c}
\hline
\multirow{2}{*}{\textbf{MLLMs}} & \multicolumn{2}{c|}
{\textbf{\textit{High-level}}} & \multicolumn{4}{c|}{\textbf{\textit{Middle-level}}} & \multicolumn{3}{c|}{\textbf{\textit{Low-level}}} & \multirow{2}{*}{\textbf{Average}} & \multirow{2}{*}{\textbf{Rank}} \\ \cline{2-10}
 & \textit{Eye-Catch} & \textit{Comp.} & \textit{Subj. Int.} & \textit{Subj. Clut.} & \textit{Back. Clut.} & \textit{Lvl. Shot} & \textit{Img. Clarity} & \textit{Exposure} & \textit{Saturation} & & \\ 
\hline
Qwen2-VL-7B & 0.627/0.692 & 0.616/0.663 & N.A. & N.A. & 0.269/0.265 & N.A. & 0.444/0.518 & 0.171/0.191 & 0.507/0.559 & N.A. & 7 \\ 
Ours & \textbf{0.711/0.755} & \textbf{0.825/0.854} & \textbf{0.493/0.491} & \textbf{0.552/0.641} & \underline{0.380/0.380} & \textbf{0.405/0.405} & \textbf{0.640/0.702} & \textbf{0.535/0.553} & \textbf{0.752/0.774} & \textbf{0.588/0.617} & 1 \\ 
\cline{1-12} 
GPT-4o \cite{gpt4o} & \underline{0.702/0.722} & \underline{0.795/0.812} & \underline{0.232/0.231} & 0.337/0.372 & \textbf{0.416/0.412} & 0.238/0.238 & \underline{0.625/0.676} & \underline{0.520/0.513} & 0.633/0.655 & \underline{0.500/0.515} & 2 \\ 
Qwen-VL-Max \cite{qwen2vl-max} & 0.648/0.694 & 0.701/0.749 & 0.231/0.232 & \underline{0.461/0.488} & 0.381/0.376 & 0.273/0.273 & 0.581/0.622 & 0.348/0.361 & 0.683/0.686 & 0.479/0.498 & 3 \\ 
GPT-4o-mini \cite{gpt4omini} & 0.678/0.707 & 0.753/0.791 & 0.164/0.167 & 0.327/0.341 & 0.354/0.356 & 0.266/0.266 & 0.525/0.549 & 0.293/0.313 & \underline{0.736/0.763} & 0.455/0.473 & 4 \\ 
GPT-4V \cite{gpt4v} & 0.656/0.708 & 0.713/0.755 & 0.212/0.213 & 0.362/0.397 & 0.363/0.368 & \underline{0.319/0.319} & 0.500/0.576 & 0.282/0.311 & 0.585/0.600 & 0.444/0.472 & 5 \\ 
Qwen2-VL-72B  & 0.633/0.703 & 0.615/0.677 & 0.263/0.264 & 0.455/0.476 & 0.123/0.122 & 0.248/0.248 & 0.587/0.645 & 0.336/0.351 & 0.652/0.665 & 0.435/0.461 & 6 \\ 
\hline
\end{tabular}
}
\vspace{-5pt}
\caption{Evaluation results of the fine-grained attributes on the RealQA dataset. The metrics are SRCC/PLCC. }
\label{tab:mllm_res}
\vspace{-0.5cm}
\end{table*}

\paragraph{\noindent\textbf{Model Architecture.}}
As illustrated in Fig.\ref{fig:pipeline}, we utilize LoRA \cite{hu2021lora} to fine-tune Qwen2-VL-7B. Referring to Q-Align \cite{qalign}, we unfreeze the vision encoder, allowing the model to learn different levels of granularity in the input images adaptively and set the LoRA rank to 128.

\section{Experiments}
\vspace{-0.1cm}
In the following sections, we first present the dataset and implementation details in Sec.\ref{sec:dataset} and Sec.\ref{sec:imp}. Then, we demonstrate the ability of the cold-start model to extract attributes compared to general MLLMs in Sec.\ref{sec:attributes acc} and provide an ablation study on attribute prediction in Sec.\ref{sec:attrs ablation}. Next, we conduct in-depth research on using MLLMs to directly predict numerical scores in Sec.\ref{sec:q2} and compare this approach with the regression method in Sec.\ref{sec:reg}. Finally, we present the comprehensive quantitative results in Sec.\ref{sec:results}.

\subsection{Datasets and Metrics}\label{sec:dataset}
\vspace{-0.1cm}

We use six realist public datasets catering to the IAA, IQA and VQA tasks. For IAA, we adopt AVA \cite{ava} and TAD66K\cite{tad66k}. For IQA, we utilize KonIQ-10k\cite{hosu2020koniq}, SPAQ \cite{spaq} and LIVE Challenge \cite{livec}. For VQA, we use KoNViD \cite{konvid}. Although there are related IQA using synthetic datasets, they are beyond the scope of our discussion. Following the previous methods \cite{qalign,tanet}, we use the Pearson linear correlation coefficient (PLCC) and Spearman rank correlation coefficient (SRCC) to evaluate the numerical scores and the fine-grained attributes. 
To assess the fine-grained attributes, we convert qualitative attributes into quantitative values by assigning numerical codes (e.g., cluttered = 1, moderately cluttered = 2, uncluttered = 3). 
\subsection{Implementation Details}\label{sec:imp}
Based on Qwen2-VL-7B, we explore the capabilities of MLLMs in quality and aesthetic scoring. We uniformly train the model for 2 epochs. When CoT data is involved, we train for 6 epochs to achieve full convergence. We fine-tune the model with 4 NVIDIA A6000 GPUs.
\subsection{Comparison with General MLLMs} \label{sec:attributes acc}
\vspace{-0.1cm}
We evaluate the proposed method compared with general MLLMs to extract the fine-grained attributes on the RealQA dataset. The general MLLMs adopt the prompts of the annotation of the close-source MLLMs, which are carefully designed. As shown in Tab.\ref{tab:mllm_res}, the proposed method significantly outperforms competitors in both the SRCC and PLCC. For Qwen2-VL-7B, we annotate results as N.A. (i.e., not applicable) due to failures to follow certain instructions. The proposed model consistently achieves the highest average rank of 1, outperforming other competitive models, such as GPT-4o, Qwen-VL-Max, and GPT-4V. 

\subsection{Ablation of Fine-grained Attributes Training} \label{sec:attrs ablation}
To accommodate diverse requirements and improve the capability of extracting the fine-grained attributes, we split the training granularity of the fine-grained attributes into ``Attributes", ``Levels" and ``Mix" as shown in Sec.\ref{sec:training_pipeline}. We show the ablation results in Tab.\ref{tab:mixed_tag}. Referring to the first three rows, we find that aggregating attributes into a conversation can get better performance. Considering the mutual influence between different levels of attributes, we further conduct the ablation study on the order of levels using the ``Mix" mode. Predicting high-level attributes first gives similar results as predicting low-level attributes first, and we choose the former by default. Finally, it can be observed that the "Mix" mode, which combines data from different training granularities, yields the best performance.

\begin{table}[!t]
    \centering
    \resizebox{0.9\linewidth}{!}{
    \begin{tabular}{c c | c c} \hline
        \textbf{Training granularity} & \textbf{Order of levels} & \textbf{SRCC} & \textbf{PLCC} \\ \hline
        Attributes & N.A. & 0.541 & 0.571 \\ 
        Levels     & N.A. & 0.567 & 0.595\\ 
        Mix        & High$\rightarrow$ Middle$\rightarrow$ Low & 0.570 & 0.593 \\ 
        Mix        & Low$\rightarrow$ Middle$\rightarrow$ High & 0.564 & 0.592 \\ \hline
        \textbf{Mix + Levels + Attributes} & \textbf{High$\rightarrow$ Middle$\rightarrow$ Low} & \textbf{0.588} & \textbf{0.617} \\ \hline
    \end{tabular}
    }
    \caption{Ablation of the fine-grained attributes training.}
    \vspace{-0.2cm}
    \label{tab:mixed_tag}
\end{table}

\subsection{Directly Predict Numerical Scores by MLLMs} \label{sec:q2}
\begin{table}[!t]
\centering
\resizebox{0.75\linewidth}{!}{
\begin{tabular}{cccc}
\hline
\textbf{Methods}    & \textbf{Accuracy}& \textbf{Recall}& \textbf{Hallucination}\\ \hline
LLaVA-1.5-7B\cite{guo2024llava} & 0.075& 0.953& 0.020\\
mPLUG-Owl2-7B \cite{mplug-owl2} & 0.399& 0.951& 0.024\\
Qwen2-VL-2B  & 0.753& 0.981&0.004\\
MiniCPM-V2.5-8B \cite{yao2024minicpm} & 0.855 & 0.995 &0.003 \\
InternVL2-8B \cite{internvl2} & 0.957& 0.997& 0.003\\ \hline
\textbf{Qwen2-VL-7B}& \textbf{0.968}& \textbf{0.998}& \textbf{0.003}\\ \hline
\end{tabular}}
\caption{Comparison of numerical sorting capabilities.}
\vspace{-0.2cm}
\label{tab:sort}
\end{table}
In this section, we conduct a series of in-depth investigations into how to effectively predict numerical scores using MLLMs. To this end, a) first, we test zero-shot capability for numerical sorting of common 7B-sized MLLMs. b) Second, we perform an ablation study on the number of digits that should be predicted in the next token paradigm. c) Third, we monitor NCM and its variant NCM$^*$ during training to determine whether MLLMs memorizes the tokens at different positions or understands them as a whole.

\paragraph{\noindent\textbf{Zero-shot Capability for Numerical Sorting.}} We conduct a toy example to test the ability of popular open-sourced MLLMs. Specifically, we randomly generate 10 decimals with a maximum valid digit of 2 and range from 1 to 10, then we prompt MLLMs to sort them from low to high. We repeat the experiment 200 times and show the average results in Tab.\ref{tab:sort}. We take 3 metrics, which are accuracy, recall and hallucination. Accuracy represents the accuracy of the predicted sequence compared to the GT sorted sequence. Recall indicates the proportion of the predicted sequence that uses the numbers to be sorted. Hallucination indicates the proportion of the predicted sequence that uses the numbers that are not in the sequence to be sorted. Interestingly, MLLMs perform quite differently on numerical sorting. Qwen2-VL-7B achieves the best results. However, basic models with less training data, such as LLaVA-1.5-7B, perform poorly on the numerical sorting task and are almost guessing. The basic model of Q-Align (i.e., mPLUG-Owl2-7B) also obtains poor results. 

\paragraph{\noindent\textbf{Ablation of the Number of Predicted Digits.}}  Intuitively, we believe the larger the $M$ in the GT numerical score $S_{GT}$, the better the effect. This is because the numerical scores (i.e., MOS) are fine-grained labels. When the numerical score is used as a label, the quantization error is small. For example, when $M$=3 for the GT numerical score $S_{GT}$, the quantization error is $0.5\%$ of the text-defined labeling in Q-Align.
As shown in Tab.\ref{tab:digits}, if one more decimal place is predicted directly, the SRCC on the AVA dataset is 0.094 more than that of directly predicting the integer. When one more decimal place is added, the effect is only slightly improved. Thus, we use $M$=3 by default in the experiment.

\begin{table}[!t]
\centering
\resizebox{0.98\linewidth}{!}{
\begin{tabular}{ccccc}
\hline
   \textit{Test set} & \multicolumn{2}{c}{\textbf{AVA}} & \multicolumn{2}{c}{\textbf{KonIQ-10k}} \\ \hline
   \textbf{Number of digits} $\bm{M}$ & SRCC        & PLCC       & SRCC          & PLCC          \\ \hline
M=1 & 0.712       & 0.705      & 0.940         & 0.950         \\
M=2 & 0.806 {\color{ForestGreen}\scriptsize (+0.094)} & 0.805 {\color{ForestGreen}\scriptsize (+0.100)} & 0.938 {\color{ForestGreen}\scriptsize (-0.002)} & 0.948 {\color{ForestGreen}\scriptsize (-0.002)} \\
M=3 & 0.808 {\color{ForestGreen}\scriptsize (+0.096)} & 0.806 {\color{ForestGreen}\scriptsize (+0.101)} & 0.939 {\color{ForestGreen}\scriptsize (-0.001)} & 0.949 {\color{ForestGreen}\scriptsize (-0.001)} \\ \hline
\end{tabular}}
\caption{Ablation of the number of digits $M$ training with the next token prediction paradigm on the AVA and KonIQ-10k dataset.
Predicting two extra digits significantly improves SRCC and PLCC on the AVA dataset.
}
\label{tab:digits}
\end{table}

\begin{figure*}[!t] 
    \centering
    \begin{minipage}{0.60\textwidth} 
        \centering
        \includegraphics[width=\textwidth]{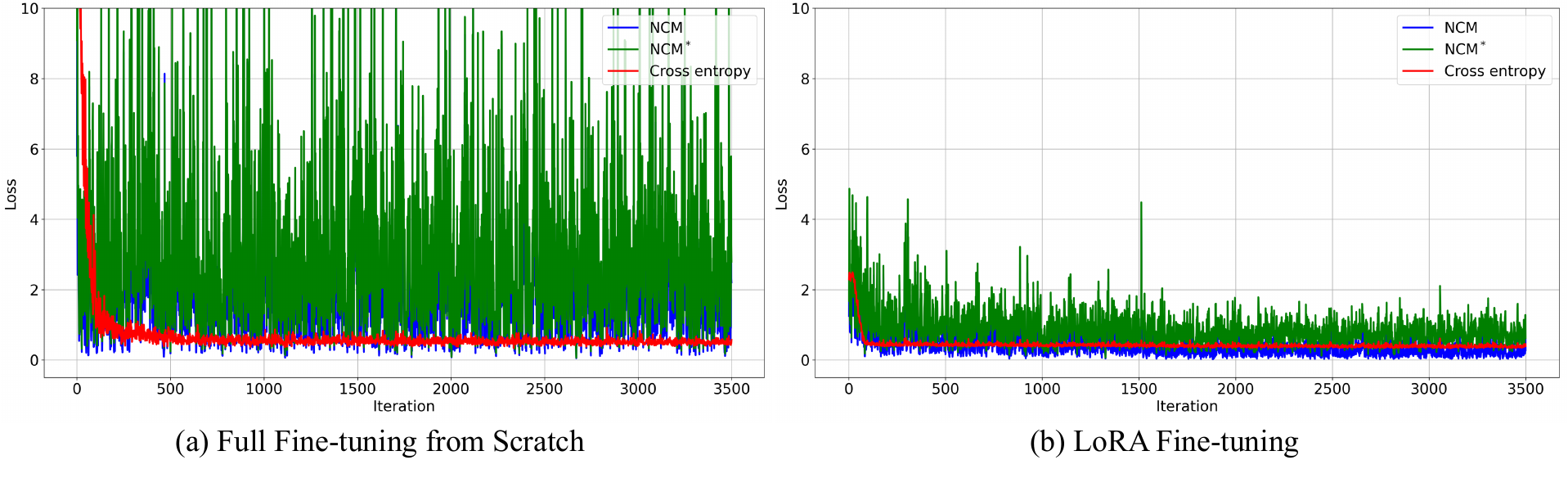}
        \caption{NCM, NCM$^*$ and $\mathcal{L}_{ce}$ in the different training stages with the supervision of $\mathcal{L}_{ce}$. 1) During the different training stages, $\mathcal{L}_{ce}$ converges. 2) Only during LoRA fine-tuning, the NCM and NCM$^{*}$ also converge. }
        \label{fig:ncm_result}
    \end{minipage}%
    \hspace{10pt}
    \begin{minipage}{0.35\textwidth}
        \centering
        \resizebox{0.98\linewidth}{!}{
            \begin{tabular}{cccc}
            \hline
            \textbf{Training Method} & \textbf{NCM} & \textbf{NCM$^*$} & $\mathcal{L}_{ce}$ \\ \hline
            \multicolumn{4}{c}{The \textbf{first} 100 iterations} \\ \hline
            Full FT from Scratch     & 2.16& 3.19& 5.91\\
            LoRA FT  & 1.43& 1.96&  1.42\\ \hline
            \multicolumn{4}{c}{The \textbf{last} 100 iterations} \\ \hline
            Full FT from Scratch     & 2.29         & 3.08                    & 0.51\\
            LoRA FT & 0.24     & 0.71                    & 0.38\\ \hline
            \multicolumn{4}{c}{\textbf{Convergence ratio (\% (last $-$ first)/ first)}} \\ \hline
    Full FT from Scratch    & {\color{red} +6.02\%} & {\color{ForestGreen} -3.45\%} & {\color{ForestGreen} -91.37\%} \\
    LoRA FT & {\color{ForestGreen}-83.22\%} & {\color{ForestGreen} -63.78\%} &  {\color{ForestGreen}-73.24\%} \\ \hline
            \end{tabular}}
        \captionof{table}{The average of the NCM, NCM$^*$ and $\mathcal{L}_{ce}$ in the different training stages. FT denotes fine-tuning. }
        \label{tab:training_methods}
    \end{minipage}
\end{figure*}

\paragraph{\noindent\textbf{NCM and Its Variant NCM$^*$ During Training.}} In this paragraph, we show two interesting findings. \textbf{\textit{Finding 1: The numerical ambiguity of the cross-entropy occurs in training a naive model.}} As shown in Fig.\ref{fig:ncm_result}(a), training Qwen2-VL-7B from scratch leads to a substantial reduction of the cross-entropy ($\mathcal{L}_{ce}$). Nevertheless, the NCM and NCM$^*$ do not converge and remain unstable. In other words, as shown in Tab.\ref{tab:training_methods}, the convergence ratio of $\mathcal{L}_{ce}$ is $-91.37\%$, while the convergence ratio of the NCM is just $+6.02\%$, which is not converged at all. It verifies that the naive model tends to ignore the intrinsic relationship between numerical scores, which only memorizes the tokens at different positions. \textbf{\textit{Finding 2: Well-trained model regularization avoids numerical ambiguity.}} As shown in Fig.\ref{fig:ncm_result}(b) and Tab.\ref{tab:training_methods}, during the LoRA FT, NCM, NCM$^*$ and $\mathcal{L}_{ce}$ converge together. 
In summary, MLLMs trained extensively on large-scale data have adequate intrinsic capabilities to accurately predict numerical scores with only the next token prediction paradigm, which indicates that MLLMs understands the numerical scores as a whole.

\subsection{Comparison with Regression Method} \label{sec:reg}
Another way to use MLLMs to directly predict numerical scores is to regress. For the regression method, the numerical score can be regressed through the linear layer. The regression method appears in the AIGC video assessment \cite{he2024videoscore}. Referring to Eq.\ref{eq:ce}, the last token of the question text contains the entire information during training, including the input image and the texts. We extract the logits of the last token and directly map them to one dimension through a linear layer. Finally, we normalize the one-dimensional feature with the sigmoid function and use MSE to supervise it. As shown in Tab.\ref{tab:reg}, we compare the next token prediction paradigm with the regression. In the same epochs, the NTP outperforms the regression method. With more training steps, the performance of the regression method can improve, but it is still not as good as the NTP. We infer that the regression method needs to re-establish the inner distribution to adjust to a different objective, so it takes longer training time to further improve. Furthermore, the NTP offers greater advantages by comparison since the output numerical scores can be used as the context of MLLMs to support other tasks and have a wider range of application scenarios.
\begin{table}[!t]
\centering
\resizebox{0.7\linewidth}{!}{
\begin{tabular}{cccc}
\hline
\textbf{Methods}    & \textbf{Epochs} & \textbf{SRCC}  & \textbf{PLCC}  \\ \hline
Regression & 2      & 0.882 & 0.890 \\
Regression & 6      & 0.908 {\color{ForestGreen}\scriptsize (+0.026)} & 0.923 {\color{ForestGreen}\scriptsize (+0.033)}\\ \hline
\textbf{NTP}       & \textbf{2}      & \textbf{0.939} {\color{ForestGreen}\scriptsize (+0.057)} & \textbf{0.949}{\color{ForestGreen}\scriptsize (+0.059)} \\ \hline
\end{tabular}}
\caption{Comparison between the next token prediction (NTP) paradigm and regression method for Qwen2-VL-7B to predict numerical scores directly using MLLMs.}
\label{tab:reg}
\end{table}

\subsection{Quantitative results} \label{sec:results}
\begin{table}[!t]
\centering
\resizebox{\linewidth}{!}{
\begin{tabular}{lcccccc}
\hline
\textit{Test set} & \multicolumn{2}{c}{\textbf{$\text{KonIQ}_{\text{test}}$}} & \multicolumn{2}{c}{\textbf{$\text{SPAQ}^{*}_{\text{test}}$}} & \multicolumn{2}{c}{\textbf{$\text{LIVE Challenge}^{*}_{\text{test}}$}} \\ \hline
\textbf{Method} & SRCC & PLCC & SRCC & PLCC & SRCC & PLCC \\
\hline 
HyperIQA (CVPR 2020) \cite{cvpr20_HyperIQA}  & 0.906 & 0.917 & 0.788 & 0.791 & 0.749 & 0.772 \\
CLIP-IQA+ (AAAI 2023) \cite{clip_iqaplus} & 0.895 & 0.909 & 0.864 & 0.866 & 0.805 & 0.832 \\
LIQE (CVPR 2023)  \cite{cvpr_liqe}  & 0.928 & 0.912 & 0.833 & 0.846 & 0.870 & 0.830 \\
DSMix (ECCV 2024) \cite{eccv_dsmix} & 0.915 & 0.925  & -& -& 0.791 & - \\
Q-Align  (ICML 2024) \cite{qalign}  & 0.940 & 0.941 & 0.887 & 0.886 & 0.860 & 0.853 \\
LoDa (CVPR 2024) \cite{cvprLoDa}   & 0.932 & 0.944 & - & - & 0.811 & - \\
QCN (CVPR 2024) \cite{cvprqcn} & 0.934 & 0.945 & - & -& 0.840 & - \\
\hline 
\textbf{Ours} & 0.939 & 0.949 & 0.878 & 0.875 & 0.878 & 0.883 \\
\textbf{Ours}$^{\dagger}$ & \textbf{0.948} & \textbf{0.959} & \textbf{0.904} & \textbf{0.906} & \textbf{0.885} & \textbf{0.898} \\
\hline
\end{tabular}
}
\caption{Results on the IQA dataset trained with the KonIQ-10k dataset. The cross-set evaluations are labeled with $^*$. $^{\dagger}$ denotes that we mix the training data organized by CoT.}
\label{tab:iqa}
\end{table}

\paragraph{\noindent\textbf{Result on IQA.}}

For the IQA task, we train the model on the KonIQ-10k\cite{hosu2020koniq} and cross-evaluate the performance on the test set of SPAQ\cite{spaq} and LIVE Challenge\cite{livec}. As shown in Tab.\ref{tab:iqa}, the proposed method consistently outperforms previous methods across all datasets. The NTP has achieved the SOTA performance, especially in the cross-dataset scenario. With the CoT training, the proposed method improves the SRCC/PLCC by $0.9\%/1.0\%$ on the $\text{KonIQ}_{\text{test}}$. In addition, the proposed method also boosts the performance on the cross-dataset. For example, our method outperforms Q-Align by achieving a $4.5\%$ higher PLCC on the cross-domain LIVE Challenge dataset. 
\paragraph{\noindent\textbf{Result on IAA.}}
\begin{table}[!t]
\centering
\resizebox{0.98\linewidth}{!}{
\begin{tabular}{lcc}
\hline
\textit{Test set} & \multicolumn{2}{c} {\textbf{$\text{AVA}_{\text{test}}$}} \\\hline
\textbf{Method} & SRCC & PLCC  \\
\hline 
HLA-GCN (CVPR 2021) \cite{HLA-GCN}  & 0.665 & 0.687\\
MUSIQ (ICCV 2021) \cite{musiq} & 0.726 & 0.738 \\
TANet (IJCAI 2022)  \cite{tanet}  & 0.758  & 0.765\\
VILA (CVPR 2023) \cite{vila} & 0.774 & 0.774 \\
Yun et al. (ECCV 2024) \cite{yun}  & 0.808  & 0.804\\
Q-Align (ICML 2024) \cite{qalign}  & 0.822  & 0.817 \\
\hline 
\textbf{Ours} & 0.809 & 0.807 \\
\textbf{Ours}$^{\dagger}$ & \textbf{0.828} & \textbf{0.825} \\\hline
\end{tabular}
\begin{tabular}{lcc}
\hline
\textit{Test set} & \multicolumn{2}{c} {\textbf{$\text{TAD66K}_{\text{test}}^{*}$}} \\\hline
\textbf{Method} & SRCC & PLCC  \\\hline
Q-Instruct \cite{Q-instruct} & 0.137 & 0.160 \\
mPLUG-Owl2 \cite{mplug-owl2} & 0.215 & 0.198 \\
VILA \cite{vila} & 0.350 & 0.372 \\
UNIAA \cite{zhou2024uniaa} & 0.411 & 0.425 \\ \hline
\textbf{Ours} & 0.410 & 0.440 \\
\textbf{Ours}$^{\dagger}$ & \textbf{0.413} & \textbf{0.444} \\\hline
\end{tabular}
}
\caption{Results on the IAA dataset trained with the AVA dataset. The cross-set evaluations are labeled with $^{*}$. $^{\dagger}$ denotes that we mix the training data organized by CoT.}
\vspace{-0.3cm}
\label{tab:iaa_ava}
\end{table}

For the IAA task, we show the comparison with the previous methods in Tab.\ref{tab:iaa_ava}. The NTP has a competitive performance on the AVA dataset. With the CoT training, the proposed method improves the SRCC/PLCC of $1.9\%/1.8\%$. For the cross-dataset TAD66K, UNIAA trains on 5 datasets including AVA, while the proposed method only trains on AVA. The proposed method also achieves a $1.9\%$ higher PLCC compared with UNIAA.
\vspace{-0.3cm}

\paragraph{\noindent\textbf{Result on VQA.}}
For the VQA task, we first test the capability of Qwen2-VL-7B itself. On the KoNViD \cite{konvid} dataset, even after prompt engineering, the SRCC of the original Qwen2-VL-7B is only 0.305. Second, considering Qwen2-VL-7B employs the 3D convolution to extract image and video features, a natural idea is to apply the image-pretained model to the video task. To this end, based on the model trained on the IQA dataset (Koniq-10k), we directly test the performance on the VQA dataset (KoNViD). In the implementation, we replace the ``image" with ``video" in the prompt. As shown in Tab.\ref{tab:vqa}, the score generated by CoT can boost the SRCC by $36.4\%$. It demonstrates that the proposed method has strong generalization for VQA.
\vspace{-0.3cm}

\paragraph{\noindent\textbf{Results on RealQA.}}
On the RealQA dataset, we follow the division of the pre-defined textual labels for Q-Align to split the composite scores.  Then, we train Q-Align and evaluate it using its evaluation method. As shown in Tab.\ref{tab:composite score}, the proposed method still outperforms Q-Align, proving that MLLMs can directly predict numerical scores effectively. 

\begin{table}[!t]
    \centering
    \resizebox{0.55\linewidth}{!}{
    \begin{tabular}{ccc}
    \hline
 \multicolumn{1}{c}{\textbf{Baseline}}& \multicolumn{1}{c}{\textbf{CoT training}} &\textbf{SRCC} \\
    \hline
    \checkmark && 0.305 \\
 \checkmark & \checkmark  &\textbf{0.669} {\color{ForestGreen}\scriptsize (+0.364)}\\    \hline
    \end{tabular}}
\caption{Zero-shot ability on the KoNViD dataset after training with the IQA dataset, Koniq-10k. Note that the training processing has no video data .}
\label{tab:vqa}
\end{table}

\begin{table}[!t]
    \centering
    \resizebox{0.5\linewidth}{!}{ 
    \begin{tabular}{lccc} 
    \hline
        \textbf{Metrics} & \textbf{Q-Align} & \textbf{Ours} & \textbf{Ours}$^{\dagger}$ \\ 
    \hline
        PLCC & 0.716 & 0.792 & \textbf{0.802} \\ 
        SRCC & 0.741 & 0.730 & \textbf{0.746} \\ 
    \hline
    \end{tabular}
    }
    \caption{Results on the RealQA dataset. $^{\dagger}$ denotes that we mix the training data organized by CoT.}
    \vspace{-0.3cm}
    \label{tab:composite score}
\end{table}

\section{Conclusion}
In this paper, we introduce RealQA, a novel dataset of 14,715 real-world images, each of which is annotated with 10 fine-grained attributes. Besides, we conduct a series of in-depth and comprehensive investigations into how to effectively predict numerical scores using MLLMs. Surprisingly, by predicting just two extra significant digits, the next token paradigm can achieve SOTA performance. With the help of CoT combined with the fine-grained attributes, the proposed method can outperform SOTA methods on five public datasets for IQA and IAA with superior interpretability and show strong
zero-shot generalization for VQA. We sincerely hope the proposed method can inspire future quality and aesthetic scoring methods based on MLLMs.

{\small
\bibliographystyle{ieee_fullname}
\bibliography{egbib}
}
\appendix

\clearpage
\renewcommand{\thefigure}{\arabic{figure}}
\renewcommand{\thetable}{\arabic{table}}
\renewcommand{\thefootnote}{\arabic{footnote}}

\setcounter{figure}{0}

\setcounter{table}{0}
\setcounter{footnote}{0}
\begin{appendices}

\section{Details of Toy Example}\label{sup_sec:toy}
The toy example aims to judge whether MLLMs can sort numerical scores. Therefore, we randomly generate 10 decimals 200 times with a maximum valid digit of 2 (e.g., 3, 3.8, 3.99) and calculate the metrics on average. We take 3 metrics, which are accuracy, recall and hallucination. Accuracy represents the accuracy of the predicted sequence compared to the GT sorted sequence. Recall indicates the proportion of the predicted sequence that uses the numbers to be sorted. Hallucination indicates the proportion of the predicted sequence that uses the numbers that are not in the sequence to be sorted. The implementation of these metrics is as shown in algorithm \ref{algorithm:metrics}. As shown in Tab.\ref{tab:sub_toy}, we show some examples of the numerical sorting experiment. It can be seen that Qwen2-VL-7B can sort the numbers correctly regardless of whether there are repeated numbers or numbers with different significant digits. Nevertheless, Qwen2-VL-7B exhibits a few instances of errors, such as hallucinated red numbers in the final line.

\begin{table*}[!t]
\centering
\resizebox{0.9\textwidth}{!}{
\begin{tabular}{p{0.6\textwidth}|p{0.4\textwidth}} \hline
\textbf{Prompt} & \textbf{Answer} \\\hline
Sort these numbers from low to high and return the result directly to me! Numbers: [5.9, 4.0, 8.88, 5.52, 3.5, 4.8, 9.2, 9.375, 7.26, 7.8]. Return format: [number 1, number 2, ..., number n] & [3.5, 4.0, 4.8, 5.52, 5.9, 7.26, 7.8, 8.88, 9.2, 9.375] \\ \hline
Sort these numbers from low to high and return the result directly to me! Numbers: [5.0, 9.1, 8.6, 1.5, 5.345, 2.0, 7.7, 8.929, 6.0, 1.48]. Return format: [number 1, number 2, ..., number n] & [1.48, 1.5, 2.0, 5.0, 5.345, 6.0, 7.7, 8.6, 8.929, 9.1]\\ \hline
Sort these numbers from low to high and return the result directly to me! Numbers: [9.0, 2.0, 2.0, 8.0, 1.51, 8.0, 3.6, 9.72, 6.0, 8.3]. Return format: [number 1, number 2, ..., number n] & [1.51, 2.0, 2.0, 3.6, 6.0, 8.0, 8.0, 8.3, {\color{red} {8.72}}, 9.0, 9.72]\\ \hline
\end{tabular}}
\caption{Examples of the numerical sorting experiment. It can be seen that Qwen2-VL-7B can sort the numbers correctly regardless of whether there are repeated numbers or numbers with different significant digits. Nevertheless, Qwen2-VL-7B exhibits a few instances of errors, such as hallucinated red numbers in the final line.}
\label{tab:sub_toy}
\end{table*}
\begin{algorithm}[!h]
\caption{Numerical Sorting Metrics}
\begin{algorithmic}[1]
\REQUIRE
    $gt$: List of ground truth numbers.  
    $pred$: List of predicted numbers.
\ENSURE
    $accuracy, recall, hallucination$

\STATE $gt\_set \gets \text{set}(gt)$
\STATE $pred\_set \gets \text{set}(pred)$

\STATE Compute the positions in $pred$ corresponding to $gt$ elements:

$indices \gets [\, pred.\text{index}(x) \;|\; x \in gt \text{ and } x \in pred \,]$
\IF{$indices$ is sorted in non-decreasing order}
    \STATE $accuracy \gets \frac{|indices|}{|gt|}$
\ELSE
    \STATE $accuracy \gets 0$
\ENDIF

\STATE  Calculate Recall as the fraction of ground truth items present in $pred$:

$recall \gets \frac{|\{ x \in gt \mid x \in pred\_set \}|}{|gt|}$

\STATE Calculate Hallucination as the fraction of predicted items not in the ground truth

$hallucination \gets \frac{|\{ x \in pred \mid x \notin gt\_set \}|}{|pred|}$

\RETURN $accuracy, recall, hallucination$
\end{algorithmic}
\label{algorithm:metrics}
\end{algorithm}

\section{More Results of RealQA}
\subsection{Discrepancy with IQA/IAA}
As shown in Tab.\ref{tab:sup composite score}, we evaluate the SOTA methods trained on the KonIQ-10k and AVA dataset separately on the test set of the RealQA dataset. The proposed method demonstrates better generalization capability than Q-Align, achieving a $2\%$ higher SRCC when trained on the IQA dataset and tested on the RealQA dataset. Furthermore, if training on the RealQA dataset, the SRCC/PLCC increases significantly. It verifies that there is a discrepancy between IQA/IAA tasks and UGC image assessment.

\begin{table}[!t]
\centering
\resizebox{0.8\linewidth}{!}{
\begin{tabular}{lcccccc}
\hline
\textit{Train set} & \multicolumn{2}{c}{\textbf{$\text{KonIQ}_{\text{train}}$}} & \multicolumn{2}{c}{\textbf{$\text{AVA}_{\text{train}}$}} & \multicolumn{2}{c}{\textbf{$\text{RealQA}_{\text{train}}$}} \\ \hline
\textbf{Method} & SRCC & PLCC & SRCC & PLCC & SRCC & PLCC \\
\hline 
Q-Align & 0.577 & 0.576 & 0.519 & \textbf{0.568} & 0.741 & 0.716 \\
Ours$^{\dagger}$ & \textbf{0.597} & \textbf{0.604} & \textbf{0.521} & 0.564 & \textbf{0.746} & \textbf{0.802} \\ 
\hline
\end{tabular}
}
\caption{\textbf{Zero-shot evaluation} of the IQA (i.e., Koniq-10k) and IAA (i.e., AVA) methods on the RealQA dataset. $^{\dagger}$ denotes that we mix the training data organized by CoT.}
\label{tab:sup composite score}
\end{table}

\subsection{Annotation Granularity}
As shown in Tab.\ref{tab:Annotation_granularity}, we show the annotation granularity of the fine-grained attributes. These fine-grained attributes include the composition score and the eye-catching score, judged against natural human language principles and scored from 1 to 10. Background clutter is assessed across three levels: cluttered, moderately cluttered, and uncluttered. Subject integrity is evaluated as incomplete, partially complete, or fully complete. The presence of a level shot (yes/no), image clarity (very blurry, moderately blurry, moderately clear, very clear), exposure (overexposed, slightly overexposed, properly exposed, slightly underexposed, underexposed), and saturation (ultra-low, low, medium, high, ultra-high) are also annotated. The fine-grained attributes provide a detailed understanding of images and assist the image quality and aesthetics assessment in a more fine-grained manner, offering interpretable explanations using natural language.
\subsection{Conversations for Attributes Training}
\begin{figure*}[!t]
\begin{center}
\includegraphics[width=0.85\textwidth]{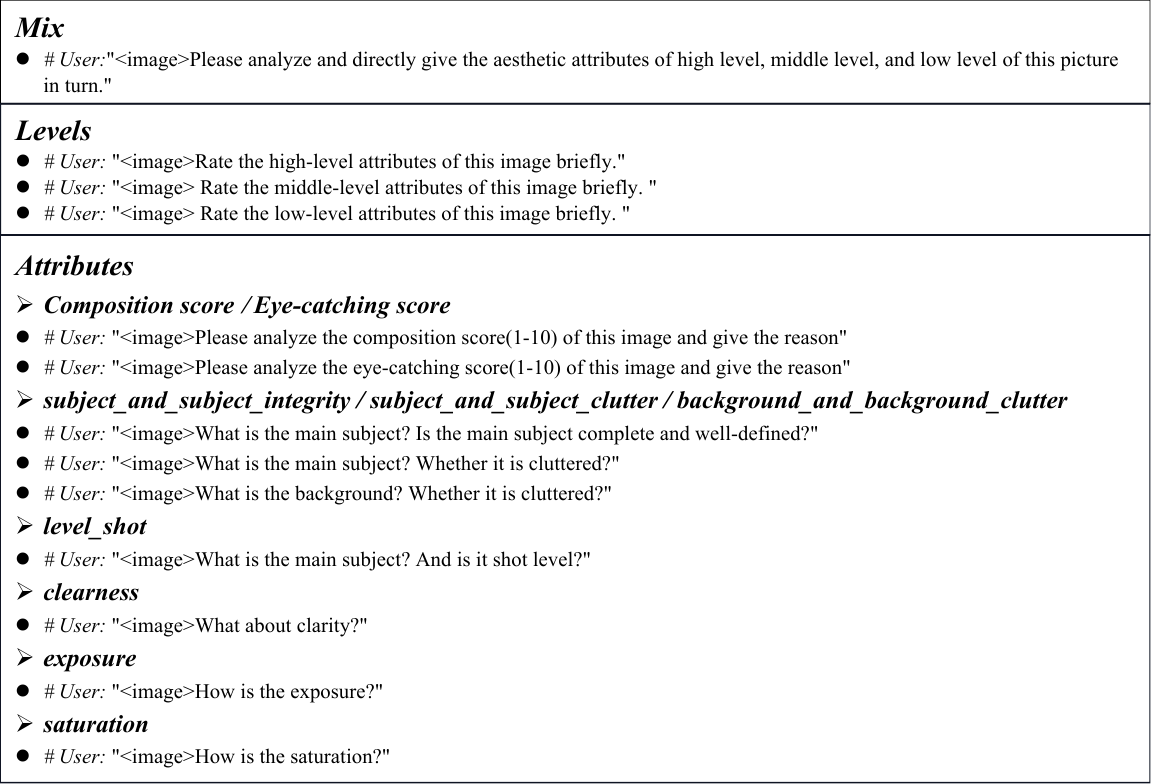}
\end{center}
   \caption{Conversations templates for the fine-grained attributes training, where the answers are omitted for ease of viewing. ``Attributes” denotes that we predict an attribute in one conversation. ``Levels” denotes that we predict certain level attributes in one conversation. ``Mix” denotes that we predict all attributes in one conversation.}
\label{fig:sup_prompt_example}

\end{figure*}

Fig.\ref{fig:sup_prompt_example} lists specific conversation templates used in the fine-grained attributes training. 

\subsection{Limitations}\label{sec:limitation}
\begin{figure*}[!t]
\begin{center}
\includegraphics[width=0.85\textwidth]{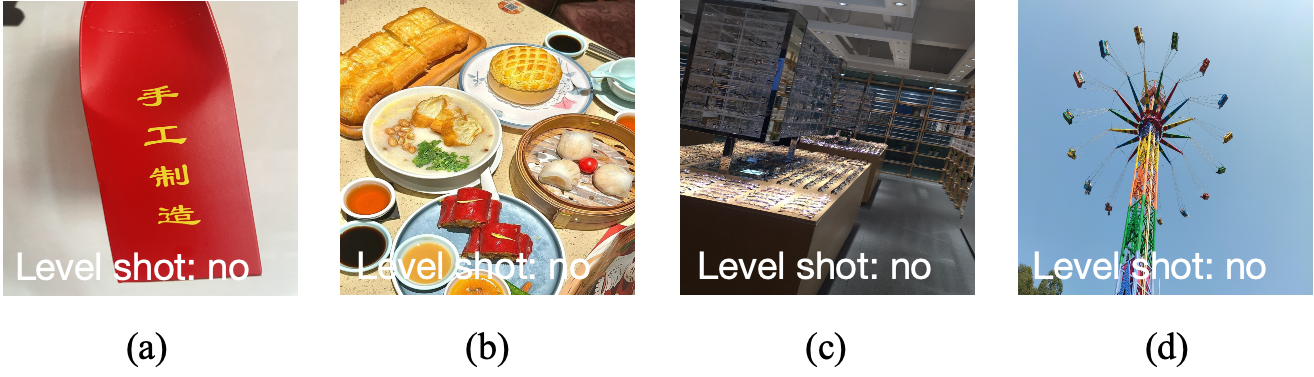}
\end{center}
   \caption{Visualization of real-world level shots. Note that all image annotations are non-level shots.  (a) It is not possible to clearly determine whether the shot is level. (b) The photo is taken from a bird's-eye view, and the dining table is tilted, making it difficult to tell whether the photo is taken level. (c) Non-level shots. (d) For entertainment facilities captured from a low angle, it is impossible to determine whether the shot is level due to the perspective. }
\label{fig:sup_shot_example}

\end{figure*}

Despite data cleaning and corrections of the results of MLLMs, we find that the proposed RealQA dataset still contains ambiguous answers. For example, judging whether an image is horizontally shot. Intuitively, this task is not difficult. However, as shown in Fig.\ref{fig:sup_shot_example}, which presents four distinct examples. It can be observed that judging whether an image is horizontally shot requires in-depth thinking about the shooting posture for the image and the relative relationship of the objects in the image to get the correct answer. In addition, we focus on common and general attributes in this work and do not separate artistic attributes. This is because artistic attributes require a certain understanding of the layout and color of the image, and even the historical stories and current politics in the image, and these data are difficult to collect and annotate. We will continue to explore the above issues in future work.

\begin{table*}[!t]
\centering
\begin{tabular}{ll}
\toprule
\textbf{Composition/Eye-catching Reason} & \textbf{Options} \\
\midrule
& natural human language \\ \bottomrule
\textbf{Composition/Eye-catching Score} & \textbf{Options} \\
\midrule
& 1-10 \\
\bottomrule
\textbf{Subject/Background Clutter} & \textbf{Options} \\
\midrule
& cluttered \\
& moderately cluttered \\
& uncluttered \\
\bottomrule
\textbf{Subject Integrity} & \textbf{Options} \\
\midrule
& incomplete (mostly cut off or mostly obscured) \\
& partially complete (slightly cut off or slightly obscured) \\
& fully complete (no cuts or obstructions at all) \\
\bottomrule
\textbf{Level Shot} & \textbf{Options} \\
\midrule
& yes \\
& no \\
\bottomrule
\textbf{Image Clarity} & \textbf{Options} \\
\midrule
& very blurry \\
& moderately blurry \\
& moderately clear \\
& very clear \\
\bottomrule
\textbf{Exposure} & \textbf{Options} \\
\midrule
& overexposed \\
& slightly overexposed \\
& properly exposed \\
& slightly underexposed \\
& underexposed \\
\bottomrule
\textbf{Saturation} & \textbf{Options} \\
\midrule
& ultra-low saturation \\
& low saturation \\
& medium saturation \\
& high saturation \\
& ultra-high saturation \\
\bottomrule
\end{tabular}
    \caption{Annotation granularity of the fine-grained attributes.}
    \label{tab:Annotation_granularity}
\end{table*}


\section{Discussion}
\subsection{Numerical Scoring for MLLMs}
Recently, there have been several hot discussions in the community: \textit{13.11 and 13.8, which one is larger}\footnote{\url{https://x.com/billyuchenlin/status/1812948314360541302?lang=en}}? Even though LLMs are capable of solving Olympiad-level mathematical problems, they still commit errors when tackling similar types of questions\footnote{\url{https://community.openai.com/t/why-9-11-is-larger-than-9-9-incredible/869824/6}}. \textbf{Please note that as of March 7, 2025, the community has not reached a consensus on why this phenomenon occurs.}
In this paper, we conduct numerical sorting on the common 7B-sized MLLMs, and the results are surprising. Although these MLLMs perform well on more complex tasks, such as VQA and GQA, their capabilities vary greatly on simple numerical sorting. The results are similar to the phenomenon discussed in the community. Although the underlying principles and explanations are beyond the scope of this paper, the proposed method can provide support for subsequent MLLMs to perform quality and aesthetic scoring tasks: it is completely feasible to predict numerical scores on the well-trained MLLMs directly.

\subsection{NCM for Qwen2-VL}
MLLMs employ a variety of tokenization methods, among which Byte Pair Encoding (BPE) is a common method widely adopted. Even when the same BPE algorithm is applied, different MLLMs can produce varying tokenization results. This discrepancy arises from factors such as the BPE training corpus, vocabulary size, and other influencing elements. In this paper, we design NCM and its variant NCM$^{*}$ through the tokenizer of Qwen2-VL, and monitor the ability of MLLMs to understand numbers during the training for numerical prediction. We can verify whether MLLMs memorizes the tokens at different positions or understands them as a whole. For different MLLMs, the organization of numbers will change due to their different tokenization methods. For example, ``3.99" may be split into ``3", ``.9", ``9", which do not appear in Qwen2-VL. The designed NCM and its variant NCM$^{*}$ need to adapt to the new tokenizer method, which means that Eq. 3 and Eq. 6 in the main paper for calculating the mathematical expectation need to be adjusted accordingly. However, we emphasize that \textbf{despite the changes in implementation details, the core ideas and theoretical basis of the proposed NCM and its variant NCM$^{*}$ remain unchanged.}

\section{Visualization Results}
\begin{figure*}[!t]
\begin{center}
\includegraphics[width=0.95\textwidth]{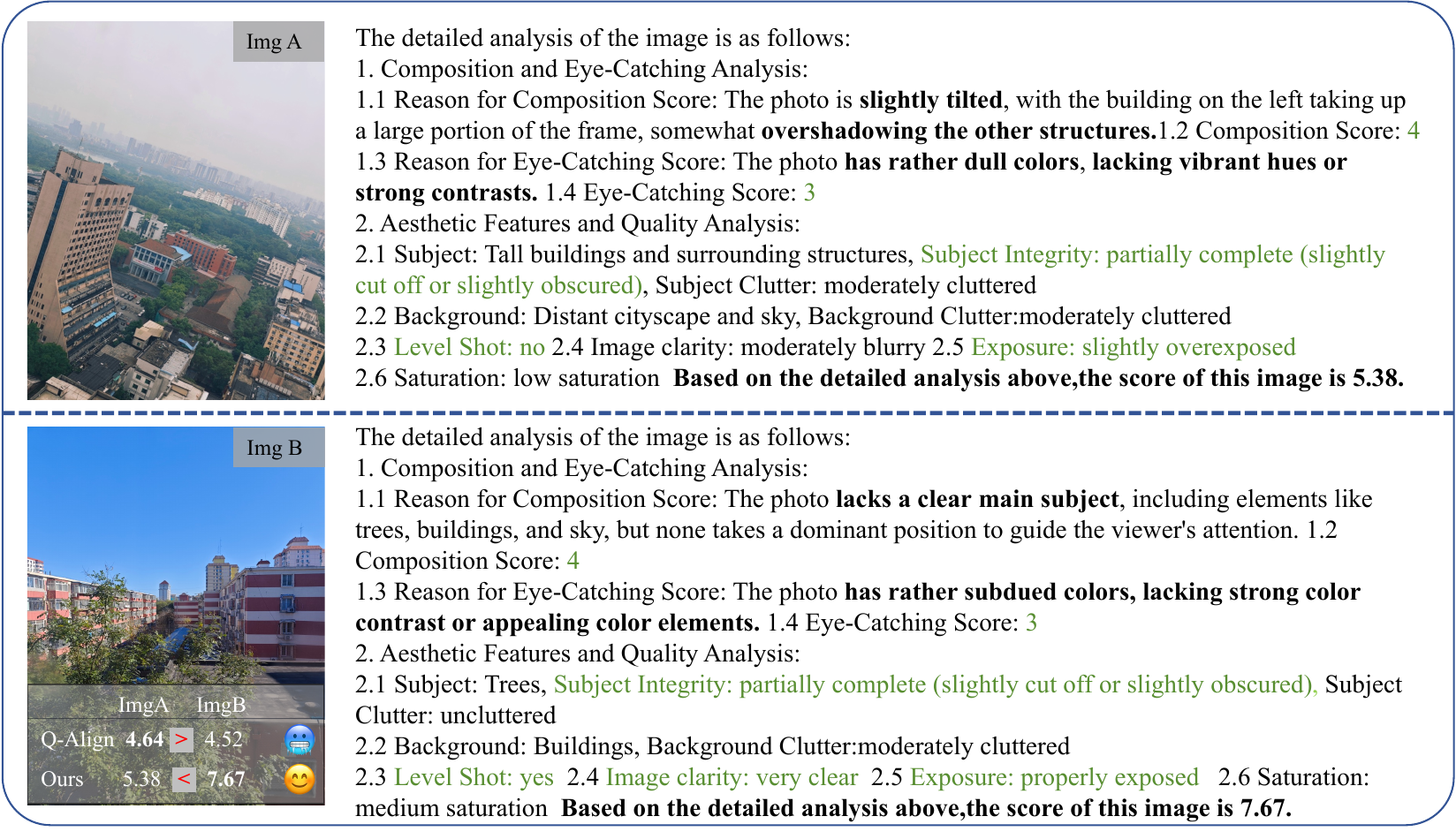}
\end{center}
   \caption{Real-world visualization results. We scale the predicted scores of Q-Align to a range of 1-10 for comparison. }
\label{fig:sup_res_1}
\end{figure*}

\begin{figure*}[!t]
\begin{center}
\includegraphics[width=0.95\textwidth]{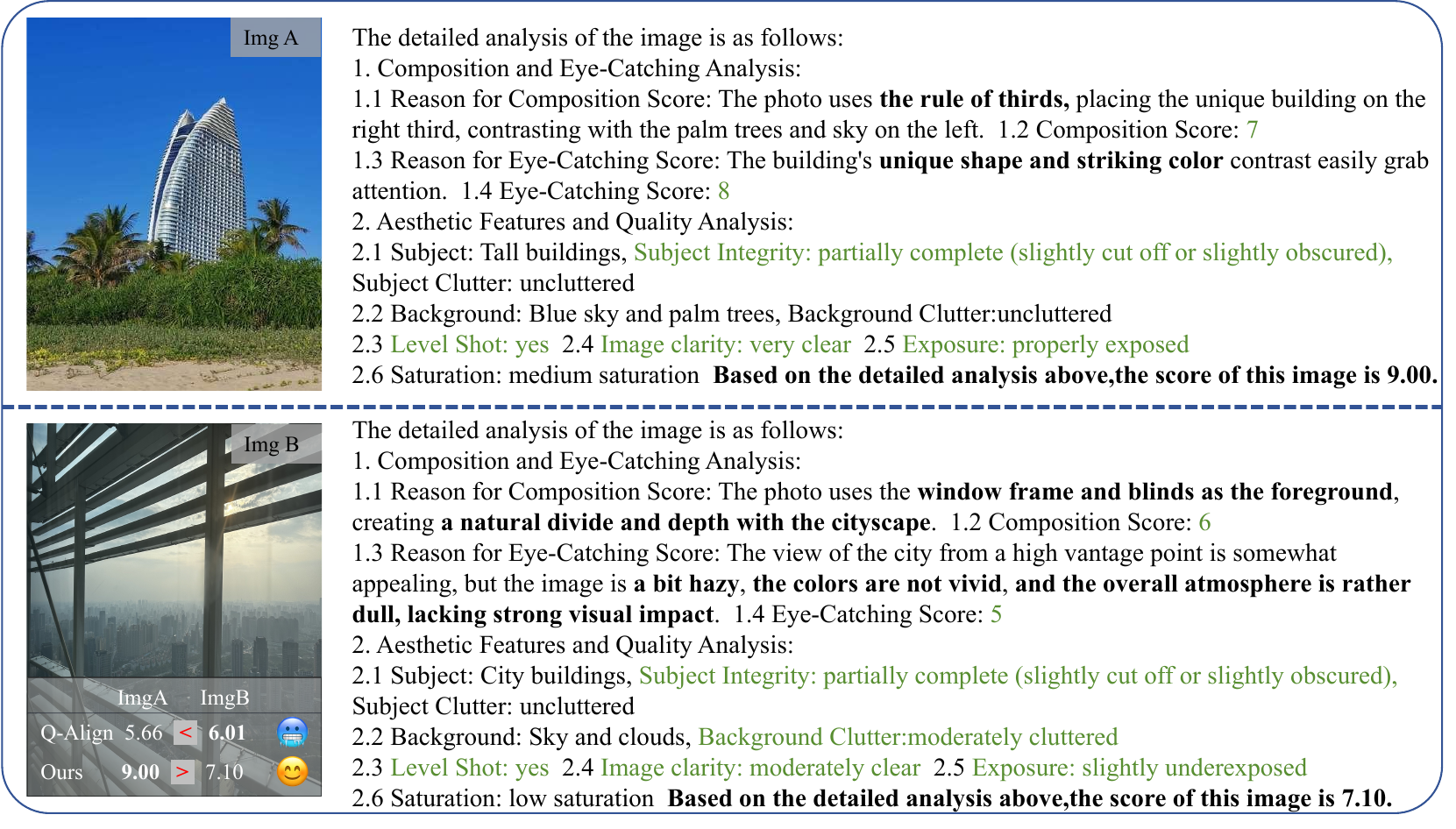}
\end{center}
   \caption{Real-world visualization results. We scale the predicted scores of Q-Align to a range of 1-10 for comparison.}
\label{fig:sup_res_2}
\end{figure*}

\begin{figure*}[!t]
\begin{center}
\includegraphics[width=0.95\textwidth]{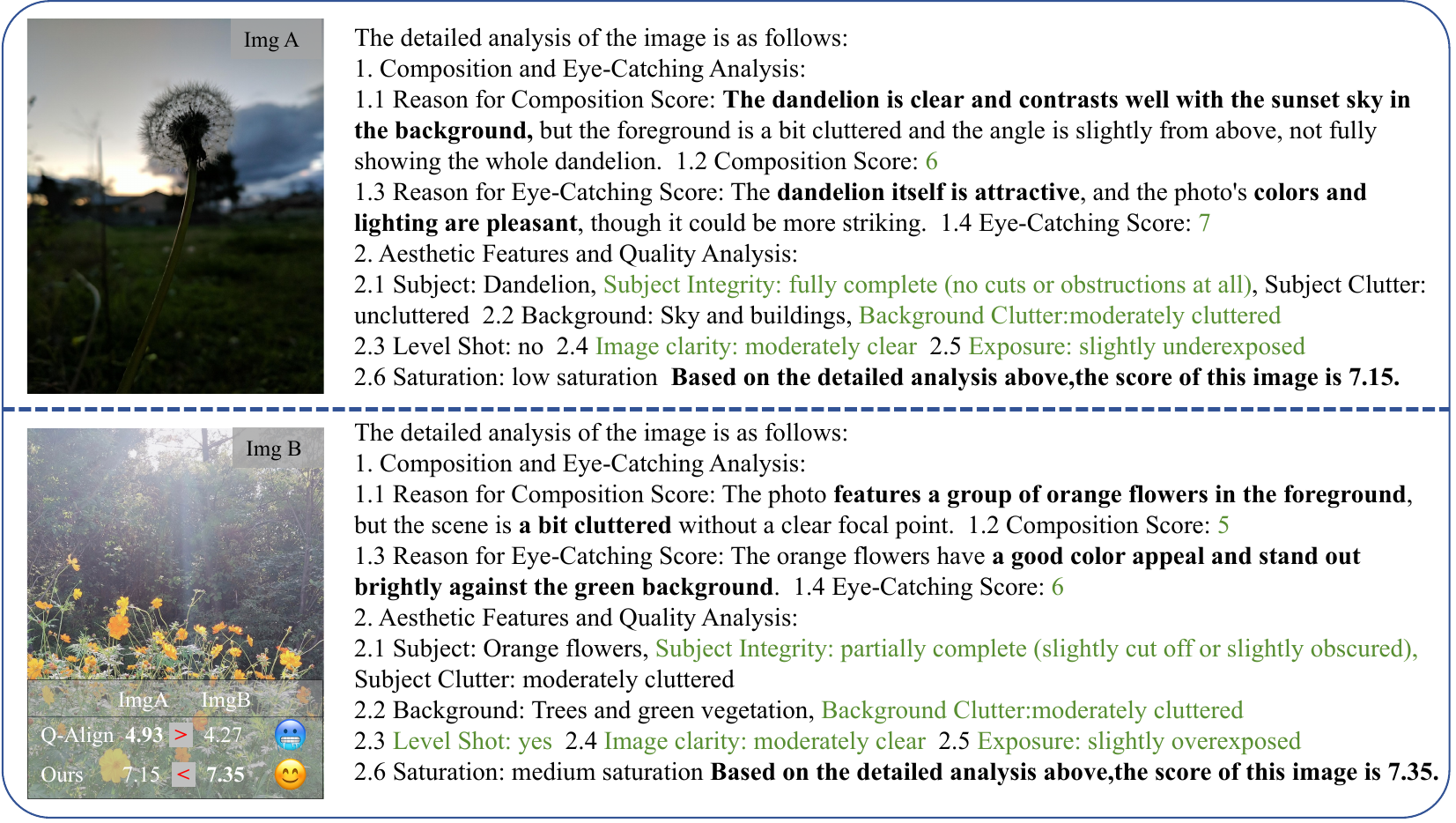}
\end{center}
   \caption{Real-world visualization results. We scale the predicted scores of Q-Align to a range of 1-10 for comparison.}
\label{fig:sup_res_3}
\end{figure*}

\begin{figure*}[!t]
\begin{center}
\includegraphics[width=0.95\textwidth]{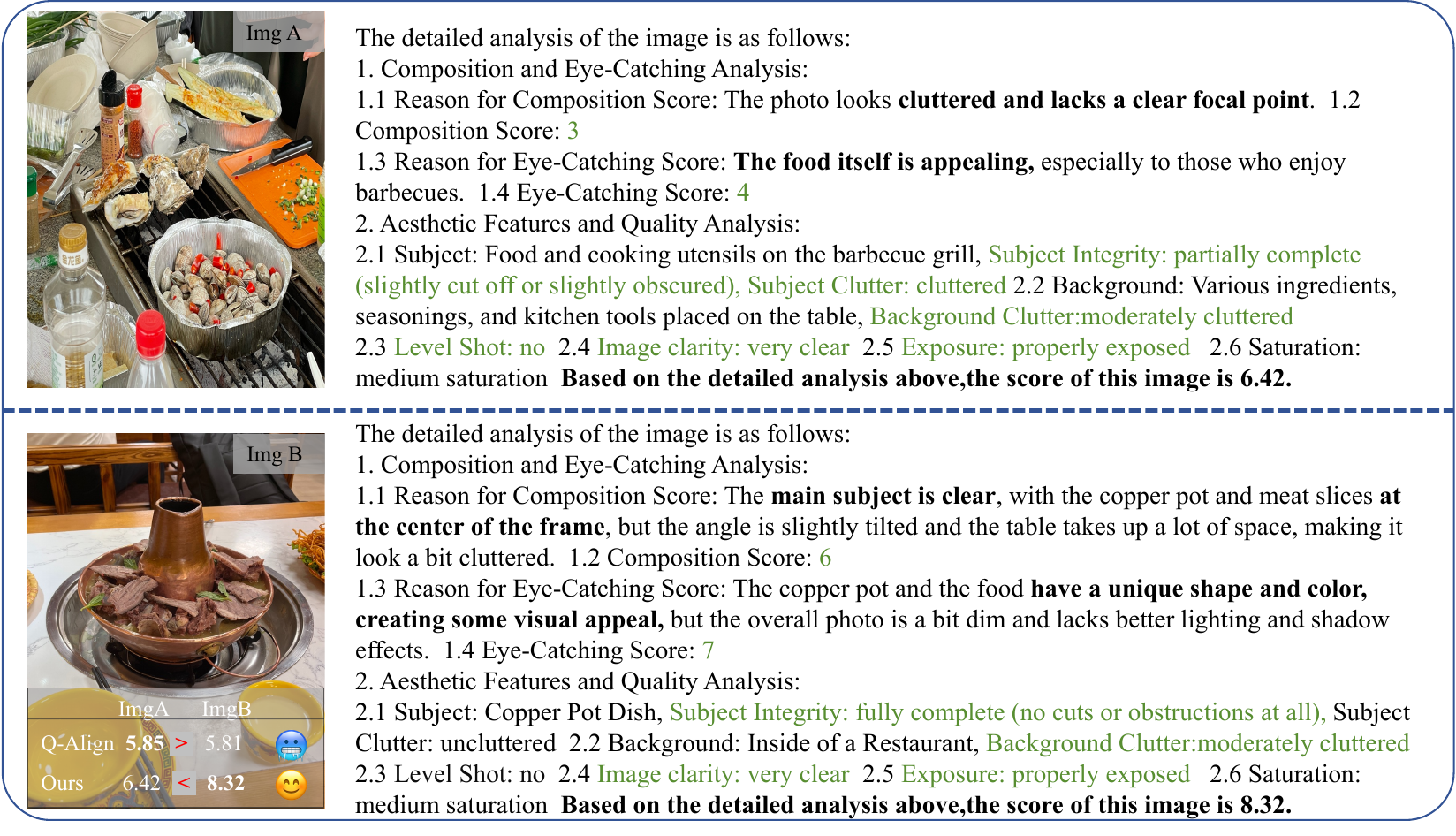}
\end{center}
   \caption{Real-world visualization results. We scale the predicted scores of Q-Align to a range of 1-10 for comparison.}
\label{fig:sup_res_4}
\end{figure*}

\begin{figure*}[!t]
\begin{center}
\includegraphics[width=0.95\textwidth]{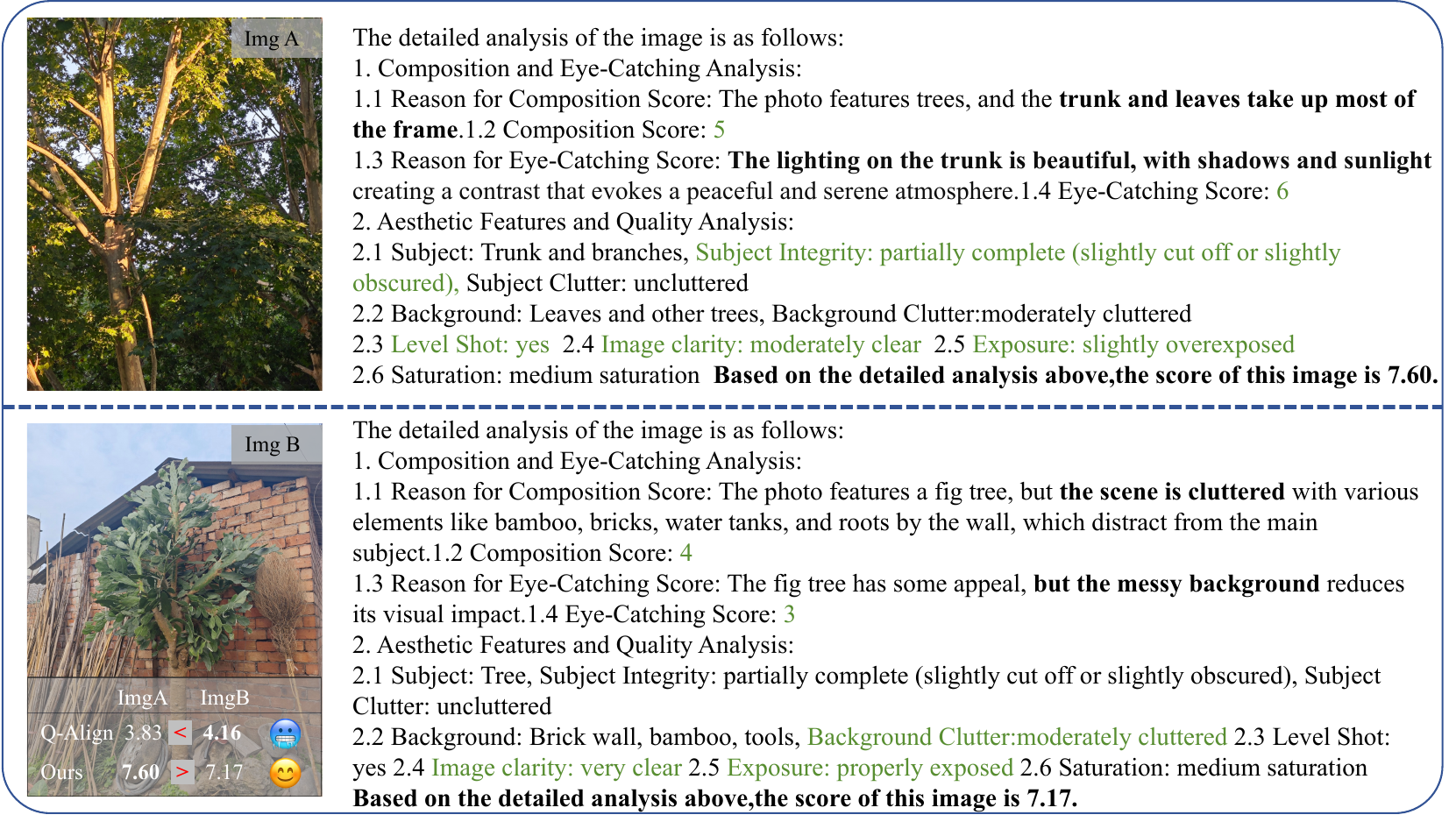}
\end{center}
   \caption{Real-world visualization results. We scale the predicted scores of Q-Align to a range of 1-10 for comparison.}
\label{fig:sup_res_5}
\end{figure*}

\begin{figure*}[!t]
\begin{center}
\includegraphics[width=0.95\textwidth]{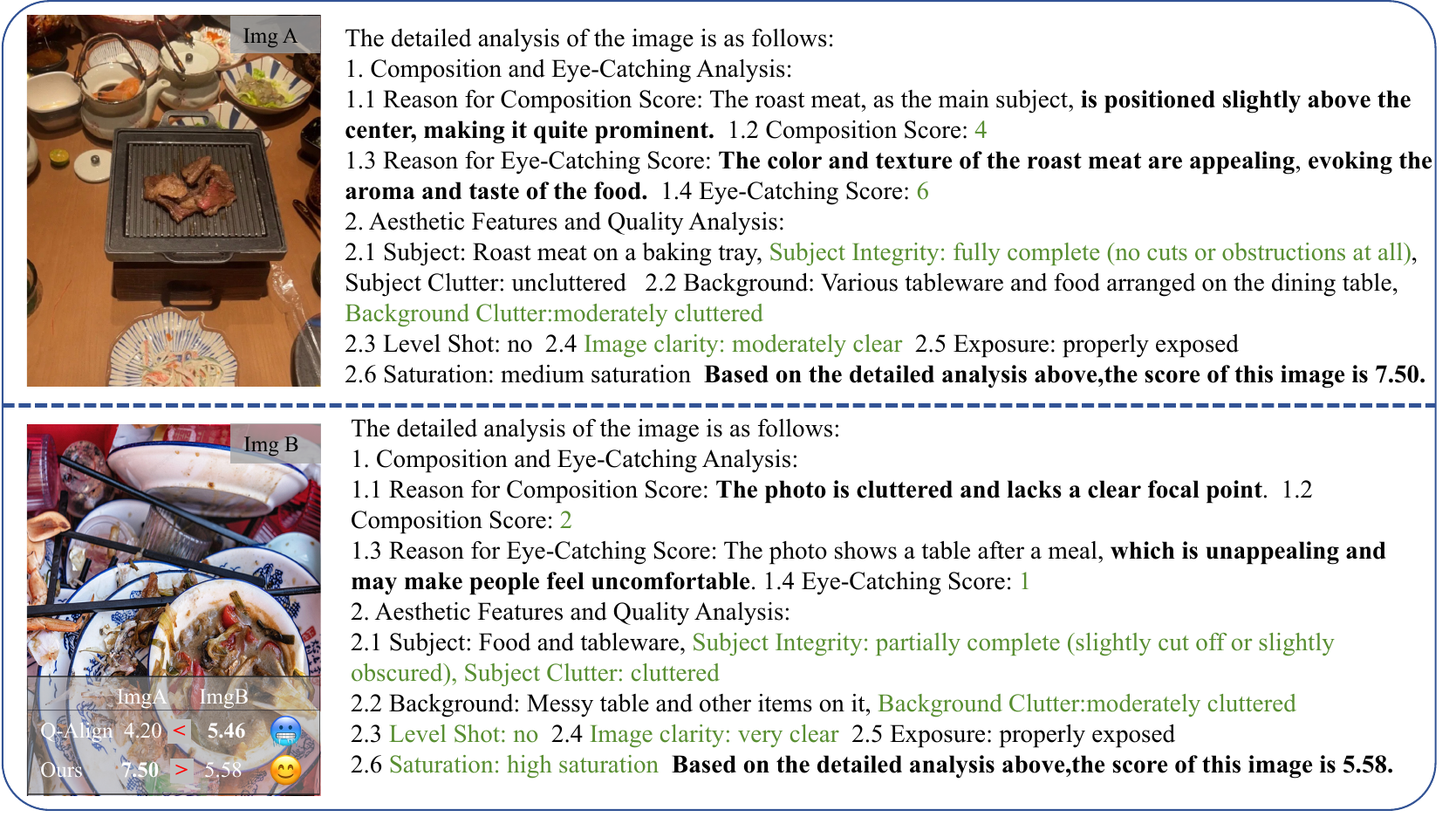}
\end{center}
   \caption{Real-world visualization results. We scale the predicted scores of Q-Align to a range of 1-10 for comparison.}
\label{fig:sup_res_6}
\end{figure*}

As shown in Fig. \ref{fig:sup_res_1} to Fig. \ref{fig:sup_res_6}, we compare the visualization results between the SOTA method (Q-Align) trained on the AVA dataset and the proposed method trained on the RealQA dataset. For images taken by daily users, the proposed method will have better ranking results of quality and aesthetics.

\end{appendices}

\end{document}